\documentclass[runningheads]{llncs}
\usepackage[T1]{fontenc}
\usepackage{graphicx}
\usepackage{ulem}
\usepackage{xcolor}
\usepackage{multirow}
\usepackage{array}
\usepackage{makecell}

\begin{document}

\title{Optimized Table Tokenization for Table Structure Recognition}

\author{Maksym Lysak\orcidID{0000-0002-3723-6960}  \and
Ahmed Nassar\orcidID{0000-0002-9468-0822} \and
Nikolaos Livathinos\orcidID{0000-0001-8513-3491} \and
Christoph Auer\orcidID{0000-0001-5761-0422} \and
Peter Staar\orcidID{0000-0002-8088-0823}}

\authorrunning{M. Lysak, et al.}
% First names are abbreviated in the running head.
% If there are more than two authors, 'et al.' is used.
%
\institute{IBM Research\\
\email{\{mly,ahn,nli,cau,taa\}@zurich.ibm.com}}
\maketitle              % typeset the header of the contribution
\begin{abstract}

% Actual limit: 1200 characters.
Extracting tables from documents is a crucial task in any document conversion pipeline. Recently, transformer-based models have demonstrated that table-structure can be recognized with impressive accuracy using Image-to-Markup-Sequence (Im2Seq) approaches. Taking only the image of a table, such models predict a sequence of tokens (e.g. in HTML, LaTeX) which represent the structure of the table. Since the token representation of the table structure has a significant impact on the accuracy and run-time performance of any Im2Seq model, we investigate in this paper how table-structure representation can be optimised. We propose a new, optimised table-structure language (OTSL) with a minimized vocabulary and specific rules. The benefits of OTSL are that it reduces the number of tokens to 5 (HTML needs 28+) and shortens the sequence length to half of HTML on average. Consequently, model accuracy improves significantly, inference time is halved compared to HTML-based models, and the predicted table structures are always syntactically correct. This in turn eliminates most post-processing needs. Popular table structure data-sets will be published in OTSL format to the community.

\end{abstract}

\section{Introduction}
Tables are ubiquitous in documents such as scientific papers, patents, reports, manuals, specification sheets or marketing material. They often encode highly valuable information and therefore need to be extracted with high accuracy. Unfortunately, tables appear in documents in various sizes, styling and structure, making it difficult to recover their correct structure with simple analytical methods. Therefore, accurate table extraction is achieved these days with machine-learning based methods.

\begin{figure*}[t]
\centering
\caption{\label{OTSLvHTML_show}Comparison between HTML and OTSL table structure representation: (A) table-example with complex row and column headers, including a 2D empty span, (B) minimal graphical representation of table structure using rectangular layout, (C) HTML representation, (D) OTSL representation. This example demonstrates many of the key-features of OTSL, namely its reduced vocabulary size (12 versus 5 in this case), its reduced sequence length (55 versus 30) and a enhanced internal structure (variable token sequence length per row in HTML versus a fixed length of rows in OTSL). }
\vspace*{.3cm}
\includegraphics[width=11cm]{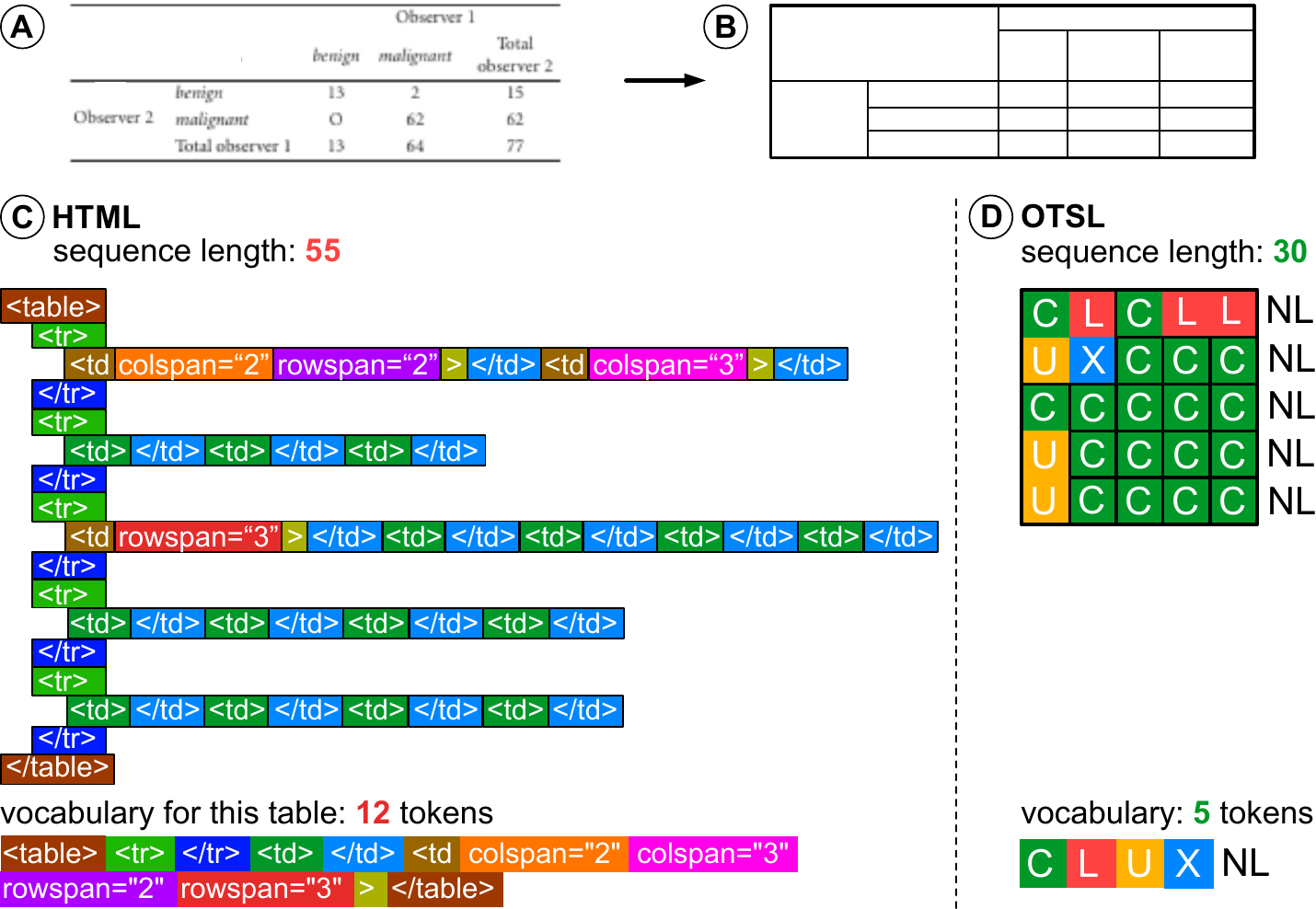}
\end{figure*}

In modern document understanding systems~\cite{IEECloud22,KDD18}, table extraction is typically a two-step process. Firstly, every table on a page is located with a bounding box, and secondly, their logical row and column structure is recognized. As of today, \textit{table detection} in documents is a well understood problem, and the latest state-of-the-art (SOTA) object detection methods provide an accuracy comparable to human observers~\cite{li2019tablebank,pagemodelrnn,DocLayNet,smock2022pubtables,PubLayNet}. On the other hand, the problem of table structure recognition (TSR) is a lot more challenging and remains a very active area of research, in which many novel machine learning algorithms are being explored~\cite{chi2019complicated,deng2019challenges,kayal2022tables,TableFormer,prasad2020cascadetabnet,schreiber2017deepdesrt,8978137,smock2022pubtables,xue2019res2tim,xue2021tgrnet,GTE,PubTabNet}.  

Recently emerging SOTA methods for table structure recognition employ transformer-based models, in which an image of the table is provided to the network in order to predict the structure of the table as a sequence of tokens. These image-to-sequence (Im2Seq) models are extremely powerful, since they allow for a purely data-driven solution. The tokens of the sequence typically belong to a markup language such as HTML, Latex or Markdown, which allow to describe table structure as rows, columns and spanning cells in various configurations. In Figure~\ref{OTSLvHTML_show}, we illustrate how HTML is used to represent the table-structure of a particular example table. Public table-structure data sets such as PubTabNet~\cite{PubTabNet}, and FinTabNet~\cite{GTE}, which were created in a semi-automated way from paired PDF and HTML sources (e.g. PubMed Central), popularized primarily the use of HTML as ground-truth representation format for TSR.

While the majority of research in TSR is currently focused on the development and application of novel neural model architectures, the table structure representation language (e.g. HTML in PubTabNet and FinTabNet) is usually adopted \textit{as is} for the sequence tokenization in Im2Seq models. In this paper, we aim for the opposite and investigate the impact of the table structure representation language with an otherwise unmodified Im2Seq transformer-based architecture. Since the current state-of-the-art Im2Seq model is TableFormer~\cite{TableFormer}, we select this model to perform our experiments.

The main contribution of this paper is the introduction of a new optimised table structure language (OTSL), specifically designed to describe table-structure in an compact and structured way for Im2Seq models. OTSL has a number of key features, which make it very attractive to use in Im2Seq models. Specifically, compared to other languages such as HTML, OTSL has a minimized vocabulary which yields short sequence length, strong inherent structure (e.g. strict rectangular layout) and a strict syntax with rules that only look backwards. The latter allows for syntax validation during inference and ensures a syntactically correct table-structure. These OTSL features are illustrated in Figure~\ref{OTSLvHTML_show}, in comparison to HTML. 

The paper is structured as follows. In section~\ref{RW}, we give an overview of the latest developments in table-structure reconstruction. In section~\ref{PS} we review the current HTML table encoding (popularised by PubTabNet and FinTabNet) and discuss its flaws. Subsequently, we introduce OTSL in section~\ref{OTSL}, which includes the language definition, syntax rules and error-correction procedures. In section~\ref{Results}, we apply OTSL on the TableFormer architecture, compare it to TableFormer models trained on HTML and ultimately demonstrate the advantages of using OTSL. Finally, in section~\ref{Con} we conclude our work and outline next potential steps.  

\section{\label{RW}Related Work}

Approaches to formalize the logical structure and layout of tables in electronic documents date back more than two decades \cite{10.5555/923400}. In the recent past, a wide variety of computer vision methods have been explored to tackle the problem of table structure recognition, i.e. the correct identification of columns, rows and spanning cells in a given table. Broadly speaking, the current deep-learning based approaches fall into three categories: object detection (OD) methods, Graph-Neural-Network (GNN) methods and Image-to-Markup-Sequence (Im2Seq) methods. Object-detection based methods \cite{prasad2020cascadetabnet,schreiber2017deepdesrt,8978137,smock2022pubtables,GTE} rely on table-structure annotation using (overlapping) bounding boxes for training, and produce bounding-box predictions to define table cells, rows, and columns on a table image. 
Graph Neural Network (GNN) based methods \cite{chi2019complicated,lee2022table,xue2019res2tim,xue2021tgrnet}, as the name suggests, represent tables as graph structures. The graph nodes represent the content of each table cell, an embedding vector from the table image, or geometric coordinates of the table cell. The edges of the graph define the relationship between the nodes, e.g. if they belong to the same column, row, or table cell. Other work \cite{zhang2022split} aims at predicting a grid for each table and deciding which cells must be merged using an attention network. 
Im2Seq methods cast the problem as a sequence generation task \cite{deng2019challenges,kayal2022tables,TableFormer,PubTabNet}, and therefore need an internal table-structure representation language, which is often implemented with standard markup languages (e.g. HTML, LaTeX, Markdown). 
In theory, Im2Seq methods have a natural advantage over the OD and GNN methods by virtue of directly predicting the table-structure. As such, no post-processing or rules are needed in order to obtain the table-structure, which is necessary with OD and GNN approaches. In practice, this is not entirely true, because a predicted sequence of table-structure markup does not necessarily have to be syntactically correct. Hence, depending on the quality of the predicted sequence, some post-processing needs to be performed to ensure a syntactically valid (let alone correct) sequence. 

Within the Im2Seq method, we find several popular models, namely the encoder-dual-decoder model (EDD)~\cite{PubTabNet}, TableFormer~\cite{TableFormer}, Tabsplitter\cite{identity_matrix} and Ye et. al.~\cite{TableMaster}. EDD uses two consecutive long short-term memory (LSTM) decoders to predict a table in HTML representation. The \textit{tag decoder} predicts a sequence of HTML tags. For each decoded table cell (\textit{<td>}), the attention is passed to the \textit{cell decoder} to predict the content with an embedded OCR approach. The latter makes it susceptible to transcription errors in the cell content of the table. TableFormer address this reliance on OCR and uses two transformer decoders for HTML structure and cell bounding box prediction in an end-to-end architecture. The predicted cell bounding box is then used to extract text tokens from an originating (digital) PDF page, circumventing any need for OCR. TabSplitter \cite{identity_matrix} proposes a compact double-matrix representation of table rows and columns to do error detection and error correction of HTML structure sequences based on predictions from \cite{TableMaster}. This compact double-matrix representation can not be used directly by the Img2seq model training, so the model uses HTML as an intermediate form. Chi et. al.~\cite{deng2019challenges} introduce a data set and a baseline method using bidirectional LSTMs to predict LaTeX code. Kayal ~\cite{kayal2022tables} introduces Gated ResNet transformers to predict LaTeX code, and a separate OCR module to extract content. 

Im2Seq approaches have shown to be well-suited for the TSR task and allow a full end-to-end network design that can output the final table structure without pre- or post-processing logic. Furthermore, Im2Seq models have demonstrated to deliver state-of-the-art prediction accuracy~\cite{TableFormer}. This motivated the authors to investigate if the performance (both in accuracy and inference time) can be further improved by optimising the table structure representation language. We believe this is a necessary step before further improving neural network architectures for this task.

\section{\label{PS}Problem Statement} 

All known Im2Seq based models for TSR fundamentally work in similar ways. Given an image of a table, the Im2Seq model predicts the structure of the table by generating a sequence of tokens. These tokens originate from a finite vocabulary and can be interpreted as a table structure. For example, with the HTML tokens \textit{<table>}, \textit{</table>}, \textit{<tr>}, \textit{</tr>}, \textit{<td>} and \textit{</td>}, one can construct simple table structures without any spanning cells. In reality though, one needs at least 28 HTML tokens to describe the most common complex tables observed in real-world documents~\cite{GTE,PubTabNet}, due to a variety of spanning cells definitions in the HTML token vocabulary. 

\begin{figure}[h]
\centering
\caption{\label{HTML_TRL}Frequency of tokens in HTML and OTSL as they appear in PubTabNet.}
\vspace*{.1cm}
\includegraphics[width=12cm]{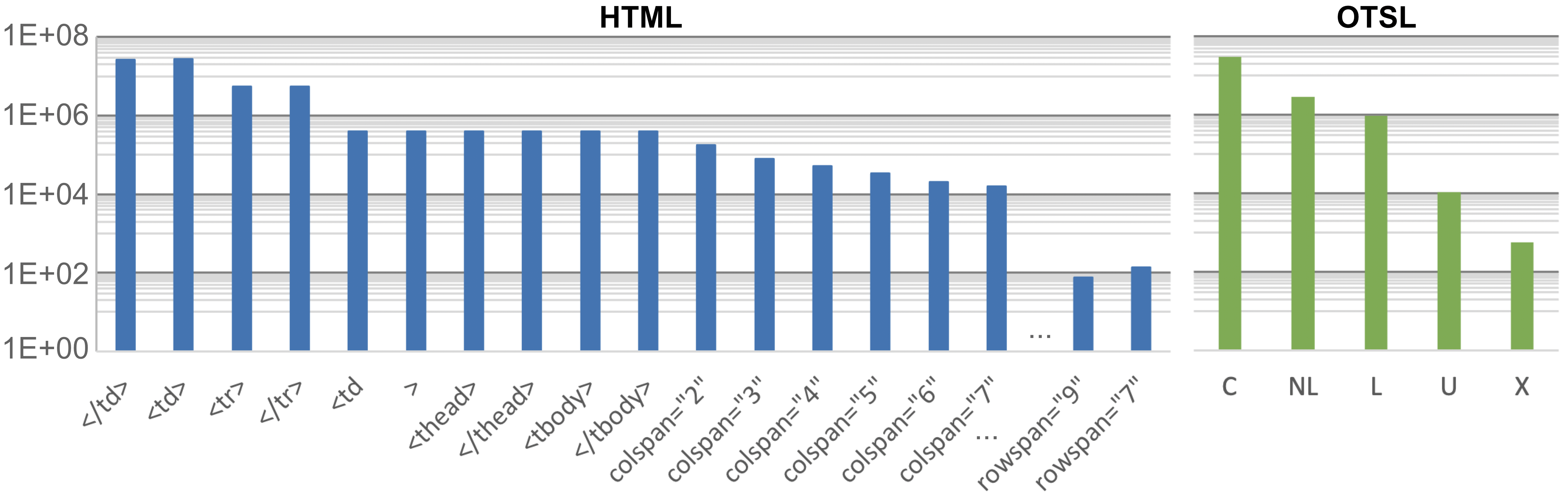}
\end{figure}

Obviously, HTML and other general-purpose markup languages were not designed for Im2Seq models. As such, they have some serious drawbacks. First, the token vocabulary needs to be artificially large in order to describe all plausible tabular structures. Since most Im2Seq models use an autoregressive approach, they generate the sequence token by token. Therefore, to reduce inference time, a shorter sequence length is critical.
Every table-cell is represented by at least two tokens (\textit{<td>} and \textit{</td>}). Furthermore, when tokenizing the HTML structure, one needs to explicitly enumerate possible column-spans and row-spans as words. In practice, this ends up requiring 28 different HTML tokens (when including column- and row-spans up to 10 cells) just to describe every table in the PubTabNet dataset. Clearly, not every token is equally represented, as is depicted in Figure~\ref{HTML_TRL}. This skewed distribution of tokens in combination with variable token row-length makes it challenging for models to learn the HTML structure.  

Additionally, it would be desirable if the representation would easily allow an early detection of invalid sequences on-the-go, before the prediction of the entire table structure is completed. HTML is not well-suited for this purpose as the verification of incomplete sequences is non-trivial or even impossible.

In a valid HTML table, the token sequence must describe a 2D grid of table cells, serialised in row-major ordering, where each row and each column have the same length (while considering row- and column-spans). Furthermore, every opening tag in HTML needs to be matched by a closing tag in a correct hierarchical manner.
Since the number of tokens for each table row and column can vary significantly, especially for large tables with many row- and column-spans, it is complex to verify the consistency of predicted structures during sequence generation. Implicitly, this also means that Im2Seq models need to learn these complex syntax rules, simply to deliver valid output.

In practice, we observe two major issues with prediction quality when training Im2Seq models on HTML table structure generation from images.
On the one hand, we find that on large tables, the visual attention of the model often starts to drift and is not accurately moving forward cell by cell anymore. This manifests itself in either in an increasing \textit{location drift} for proposed table-cells in later rows on the same column or even complete loss of vertical alignment, as illustrated in Figure~\ref{OTSLvHTML_compare}. Addressing this with post-processing is partially possible, but clearly undesired. On the other hand, we find many instances of predictions with structural inconsistencies or plain invalid HTML output, as shown in Figure~\ref{OTSLvHTML_ex3}, which are nearly impossible to properly correct.
Both problems seriously impact the TSR model performance, since they reflect not only in the task of pure structure recognition but also in the equally crucial recognition or matching of table cell content. 

\section{\label{OTSL}Optimised Table Structure Language}
To mitigate the issues with HTML in Im2Seq-based TSR models laid out before, we propose here our Optimised Table Structure Language (OTSL). OTSL is designed to express table structure with a minimized vocabulary and a simple set of rules, which are both significantly reduced compared to HTML. At the same time, OTSL enables easy error detection and correction during sequence generation. We further demonstrate how the compact structure representation and minimized sequence length improves prediction accuracy and inference time in the TableFormer architecture.

\subsection{Language Definition}
\begin{figure*}[t!]
\centering
\caption{\label{OTSL_PROOF}OTSL description of table structure: A - table example; B - graphical representation of table structure; C - mapping structure on a grid; D - OTSL structure encoding; E - explanation on cell encoding}
\vspace*{0.3cm}
\includegraphics[width=10cm]{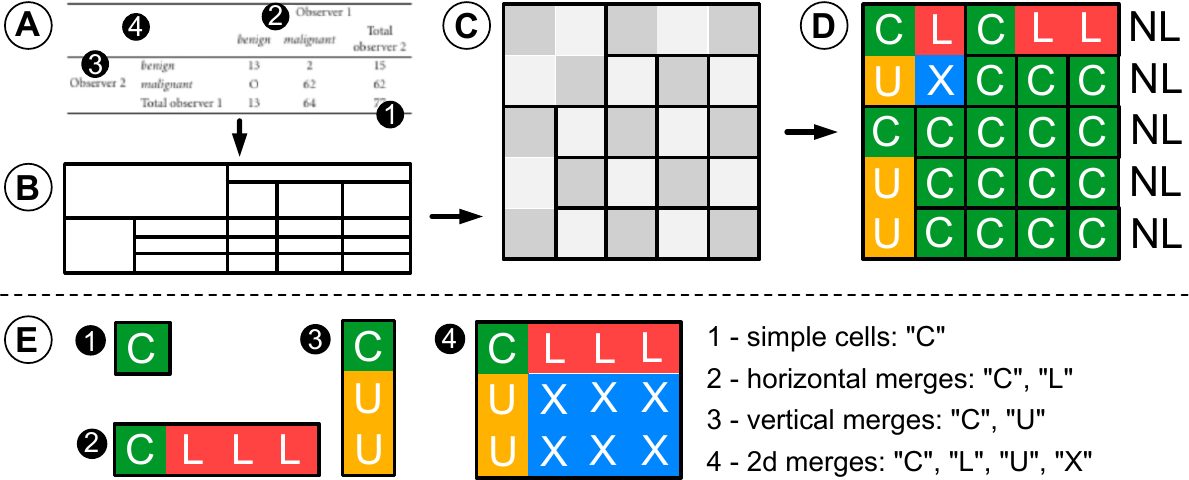}
\end{figure*}

In Figure~\ref{OTSL_PROOF}, we illustrate how the OTSL is defined. In essence, the OTSL defines only 5 tokens that directly describe a tabular structure based on an atomic 2D grid.

The OTSL vocabulary is comprised of the following tokens:
\begin{itemize}
\item "C" cell - \textit{a new table cell} that either has or does not have cell content
\item "L" cell - \textit{left-looking cell}, merging with the left neighbor cell to create a span
\item "U" cell - \textit{up-looking cell}, merging with the upper neighbor cell to create a span
\item "X" cell - \textit{cross cell}, to merge with both left and upper neighbor cells
\item "NL" - \textit{new-line}, switch to the next row.
\end{itemize}

A notable attribute of OTSL is that it has the capability of achieving lossless conversion to HTML.

\subsection{Language Syntax}

The OTSL representation follows these syntax rules:

\begin{enumerate}
\item \textbf{Left-looking cell rule}: 
The left neighbour of an "L" cell must be either another "L" cell or a "C" cell.
\item \textbf{Up-looking cell rule}:
The upper neighbour of a "U" cell must be either another "U" cell or a "C" cell.
\item \textbf{Cross cell rule}:

The left neighbour of an "X" cell must be either another "X" cell or a "U" cell, and the upper neighbour of an "X" cell must be either another "X" cell or an "L" cell.
\item \textbf{First row rule}:
Only "L" cells and "C" cells are allowed in the first row.
\item \textbf{First column rule}:
Only "U" cells and "C" cells are allowed in the first column.
\item \textbf{Rectangular rule}:
The table representation is always rectangular - all rows must have an equal number of tokens, terminated with "NL" token.
\end{enumerate}

The application of these rules gives OTSL a set of unique properties.
First of all, the OTSL enforces a strictly rectangular structure representation, where every new-line token starts a new row. As a consequence, all rows and all columns have exactly the same number of tokens, irrespective of cell spans.
Secondly, the OTSL representation is unambiguous: Every table structure is represented in one way. In this representation every table cell corresponds to a "C"-cell token, which in case of spans is always located in the top-left corner of the table cell definition.
Third, OTSL syntax rules are only backward-looking. As a consequence, every predicted token can be validated straight during sequence generation by looking at the previously predicted sequence. As such, OTSL can guarantee that every predicted sequence is syntactically valid. 

These characteristics can be easily learned by sequence generator networks, as we demonstrate further below. We find strong indications that this pattern reduces significantly the column drift seen in the HTML based models (see Figure~\ref{OTSLvHTML_compare}).

\subsection{Error-detection and -mitigation}

The design of OTSL allows to validate a table structure easily on an unfinished sequence. The detection of an invalid sequence token is a clear indication of a prediction mistake, however a valid sequence by itself does not guarantee prediction correctness. Different heuristics can be used to correct token errors in an invalid sequence and thus increase the chances for accurate predictions. Such heuristics can be applied either after the prediction of each token, or at the end on the entire predicted sequence.
For example a simple heuristic which can correct the predicted OTSL sequence on-the-fly is to verify if the token with the highest prediction confidence invalidates the predicted sequence, and replace it by the token with the next highest confidence until OTSL rules are satisfied. 

\section{\label{Results}Experiments}

\begin{figure}[h]
\centering
\caption{\label{ED_model_sketch}Architecture sketch of the TableFormer model, which is a representative for the Im2Seq approach.}
\includegraphics[width=12cm]{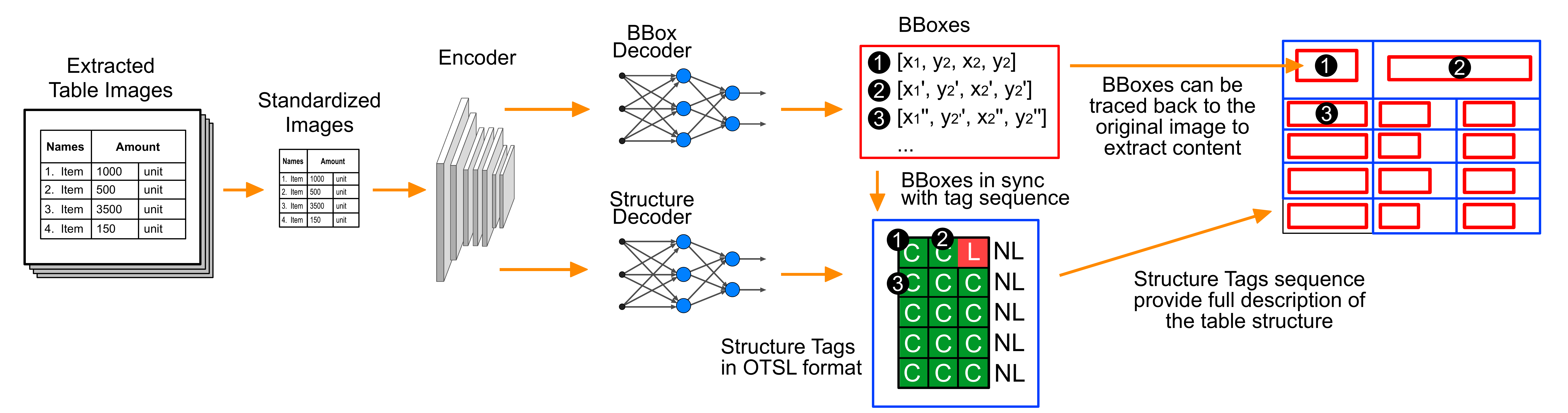}
\end{figure}

To evaluate the impact of OTSL on prediction accuracy and inference times, we conducted a series of experiments based on the TableFormer model (Figure~\ref{ED_model_sketch}) with two objectives: Firstly we evaluate the prediction quality and performance of OTSL vs. HTML after performing Hyper Parameter Optimization (HPO) on the \textit{canonical} PubTabNet data set. Secondly we pick the best hyper-parameters found in the first step and evaluate how OTSL impacts the performance of TableFormer after training on other publicly available data sets (FinTabNet, PubTables-1M~\cite{smock2022pubtables}). The ground truth (GT) from all data sets has been converted into OTSL format for this purpose, and will be made publicly available.

We rely on standard metrics such as Tree Edit Distance score (TEDs) for table structure prediction, and Mean Average Precision (mAP) with 0.75 Intersection Over Union (IOU) threshold for the bounding-box predictions of table cells.
The predicted OTSL structures were converted back to HTML format in order to compute the TED score. Inference timing results for all experiments were obtained from the same machine on a single core with AMD EPYC 7763 CPU @2.45 GHz.

\subsection{Hyper Parameter Optimization}

We have chosen the PubTabNet data set to perform HPO, since it includes a highly diverse set of tables. Also we report TED scores separately for simple and complex tables (tables with cell spans). Results are presented in Table.~\ref{tab:HPO}. It is evident that with OTSL, our model achieves the same TED score and slightly better mAP scores in comparison to HTML. However OTSL yields a \textit{2x speed up} in the inference runtime over HTML.

\setlength\extrarowheight{2pt} % or whatever amount is appropriate
\setlength{\tabcolsep}{4pt}

\begin{table}[htb!]
\centering
\caption{\label{tab:HPO} HPO performed in OTSL and HTML representation on the same transformer-based TableFormer \cite{TableFormer} architecture, trained only on PubTabNet \cite{PubTabNet}. Effects of reducing the \# of layers in encoder and decoder stages of the model show that smaller models trained on OTSL perform better, especially in recognizing complex table structures, and maintain a much higher mAP score than the HTML counterpart.}
\begin{tabular}{c|c|c|c|c|c|c|c}
\hline
\multirow{2}{*}{\begin{tabular}[c]{@{}c@{}}\#\\ enc-layers\end{tabular}} & \multirow{2}{*}{\begin{tabular}[c]{@{}c@{}}\#\\ dec-layers\end{tabular}} & \multirow{2}{*}{Language} & \multicolumn{3}{c|}{TEDs}                                        & \multirow{2}{*}{\makecell{mAP \\ (0.75)}} & \multirow{2}{*}{\makecell{Inference \\ time (secs)}} \\ \cline{4-6}
                          &                                                                          &                                                                          & \multicolumn{1}{c|}{simple} & \multicolumn{1}{c|}{complex} & all &                             &                                        \\ \hline

\multirow{2}{*}{6} & \multirow{2}{*}{6} & OTSL & 0.965  & 0.934   & 0.955  & \textbf{0.88}      & \textbf{2.73} \\
                   &                    & HTML & 0.969  & 0.927   & 0.955  & 0.857     & 5.39 \\\hline

\multirow{2}{*}{4} & \multirow{2}{*}{4} & OTSL & 0.938  & 0.904   & 0.927  & \textbf{0.853}     & \textbf{1.97} \\
                   &                    & HTML & 0.952  & 0.909   & \textbf{0.938}  & 0.843     & 3.77 \\\hline

\multirow{2}{*}{2} & \multirow{2}{*}{4} & OTSL & 0.923  & 0.897   & 0.915  & \textbf{0.859}     & \textbf{1.91} \\
                   &                    & HTML & 0.945  & 0.901   & \textbf{0.931}  & 0.834     & 3.81 \\\hline

\multirow{2}{*}{4} & \multirow{2}{*}{2} & OTSL & 0.952  & 0.92    & \textbf{0.942}  & \textbf{0.857}     & \textbf{1.22} \\
                   &                    & HTML & 0.944  & 0.903   & 0.931  & 0.824     & 2 \\ \hline

\end{tabular}
\end{table}

\subsection{Quantitative Results}

We picked the model parameter configuration that produced the best prediction quality (enc=6, dec=6, heads=8) with PubTabNet alone, then independently trained and evaluated it on three publicly available data sets: PubTabNet (395k samples), FinTabNet (113k samples) and PubTables-1M (about 1M samples). Performance results are presented in Table.~\ref{tab:HTMLvOTSL_precision}. It is clearly evident that the model trained on OTSL outperforms HTML across the board, keeping high TEDs and mAP scores even on difficult financial tables (FinTabNet) that contain sparse and large tables.

\begin{table}[htb!]
\centering
\caption{\label{tab:HTMLvOTSL_precision} TSR and cell detection results compared between OTSL and HTML on the PubTabNet~\cite{PubTabNet}, FinTabNet~\cite{GTE} and PubTables-1M~\cite{smock2022pubtables} data sets using TableFormer~\cite{TableFormer} (with enc=6, dec=6, heads=8).}
\begin{tabular}{c|c|c|c|c|c|c}
\hline
\multirow{2}{*}{Data set} & \multirow{2}{*}{\makecell{Language}} & \multicolumn{3}{c|}{TEDs}                                        & \multirow{2}{*}{mAP(0.75)} & \multirow{2}{*}{\makecell{Inference \\ time (secs)}} \\ \cline{3-5}
                                &                           & \multicolumn{1}{c|}{simple} & \multicolumn{1}{c|}{complex} & all &                            &                                       \\ \hline

\multirow{2}{*}{PubTabNet}    & OTSL  & 0.965  & 0.934   & 0.955  & \textbf{0.88}  & \textbf{2.73} \\
                              & HTML  & 0.969  & 0.927   & 0.955  & 0.857          & 5.39 \\ \hline
\multirow{2}{*}{FinTabNet}    & OTSL  & 0.955  & 0.961   & \textbf{0.959}  & \textbf{0.862} & \textbf{1.85} \\
                              & HTML  & 0.917  & 0.922   & 0.92   & 0.722          & 3.26 \\ \hline
\multirow{2}{*}{PubTables-1M} & OTSL  & 0.987  & 0.964   & \textbf{0.977}  & \textbf{0.896} & \textbf{1.79} \\ 
                              & HTML  & 0.983  & 0.944   & 0.966  & 0.889          & 3.26 \\ \hline

\end{tabular}
\end{table}

Additionally, the results show that OTSL has an advantage over HTML when applied on a bigger data set like PubTables-1M and achieves significantly improved scores. Finally, OTSL achieves faster inference due to fewer decoding steps which is a result of the reduced sequence representation.

\newpage
\subsection{Qualitative Results}

\begin{figure}[htb!]
\centering
\caption{\label{OTSLvHTML_compare}The OTSL model produces more accurate bounding boxes with less overlap (E) than the HTML model (D), when predicting the structure of a sparse table (A), at twice the inference speed because of shorter sequence length (B),(C). "PMC2807444\_006\_00.png" PubTabNet.}
\vspace*{.2cm}
\includegraphics[width=10.1cm]{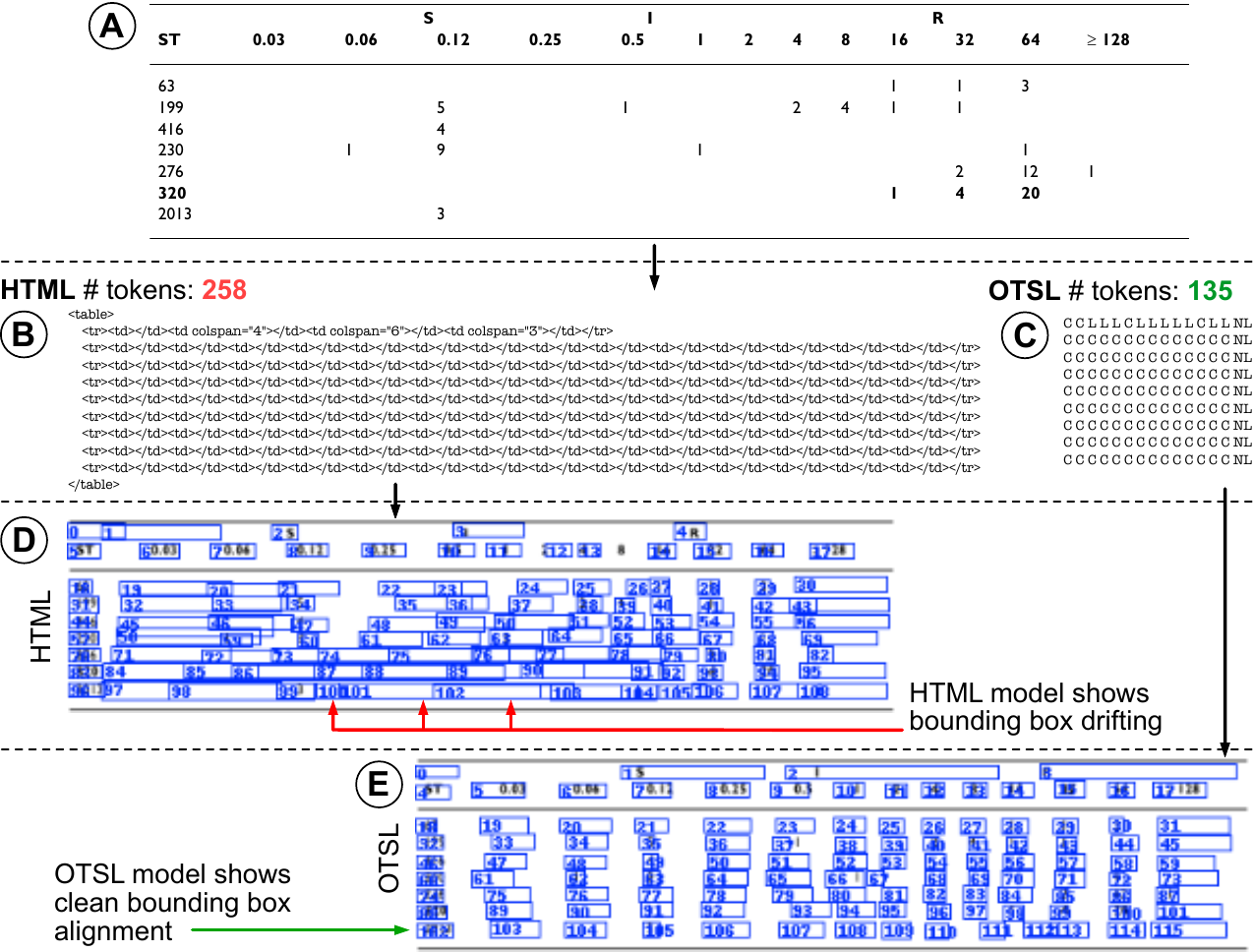}
\end{figure}

\begin{figure}[htb!]
\centering
\caption{\label{OTSLvHTML_ex3}Visualization of predicted structure and detected bounding boxes on a complex table with many rows. The OTSL model (B) captured repeating pattern of horizontally merged cells from the GT (A), unlike the HTML model (C). The HTML model also didn't complete the HTML sequence correctly and displayed a lot more of drift and overlap of bounding boxes. "PMC5406406\_003\_01.png" PubTabNet.
}
\vspace*{.2cm}
\includegraphics[width=9.8cm]{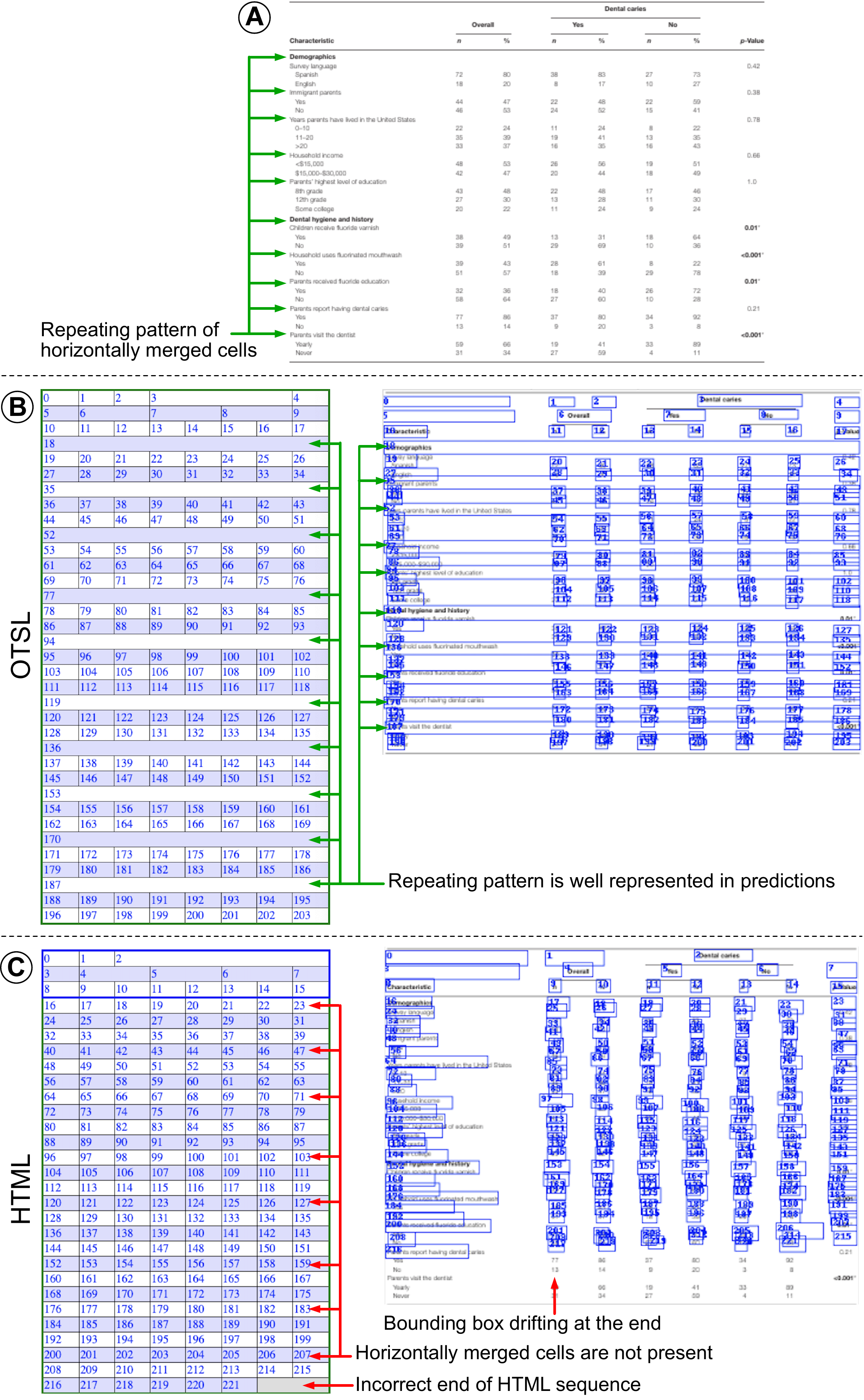}
\end{figure}

To illustrate the qualitative differences between OTSL and HTML, Figure~\ref{OTSLvHTML_compare} demonstrates less overlap and more accurate bounding boxes with OTSL. In Figure~\ref{OTSLvHTML_ex3}, OTSL proves to be more effective in handling tables with longer token sequences, resulting in even more precise structure prediction and bounding boxes.

\newpage

\newpage
\section{\label{Con}Conclusion}
We demonstrated that representing tables in HTML for the task of table structure recognition with Im2Seq models is ill-suited and has serious limitations. Furthermore, we presented in this paper an Optimized Table Structure Language (OTSL) which, when compared to commonly used general purpose languages, has several key benefits. 

First and foremost, given the same network configuration, inference time for a table-structure prediction is about 2 times faster compared to the conventional HTML approach. This is primarily owed to the shorter sequence length of the OTSL representation. Additional performance benefits can be obtained with HPO (hyper parameter optimization). As we demonstrate in our experiments, models trained on OTSL can be significantly smaller, e.g. by reducing the number of encoder and decoder layers, while preserving comparatively good prediction quality. This can further improve inference performance, yielding 5-6 times faster inference speed in OTSL with prediction quality comparable to models trained on HTML (see Table~\ref{tab:HPO}).

Secondly, OTSL has more inherent structure and a significantly restricted vocabulary size. This allows autoregressive models to perform better in the TED metric, but especially with regards to prediction accuracy of the table-cell bounding boxes (see Table~\ref{tab:HTMLvOTSL_precision}). As shown in Figure~\ref{OTSLvHTML_compare}, we observe that the OTSL drastically reduces the drift for table cell bounding boxes at high row count and in sparse tables. This leads to more accurate predictions and a significant reduction in post-processing complexity, which is an undesired necessity in HTML-based Im2Seq models.
Significant novelty lies in OTSL syntactical rules, which are few, simple and always backwards looking. Each new token can be validated only by analyzing the sequence of previous tokens, without requiring the entire sequence to detect mistakes. This in return allows to perform structural error detection and correction on-the-fly during sequence generation.

\bibliographystyle{splncs04}
\bibliography{papers}

\end{document}